\documentclass{article}

\usepackage[final]{corl_2019} % Uncomment for the camera-ready ``final'' version

\usepackage{multicol}
\usepackage{color}
\usepackage{amssymb}
\usepackage{amsmath}
\usepackage{amsthm}
\usepackage{tikz}
\usepackage{chngcntr}
\usepackage[makeroom]{cancel}
\usetikzlibrary{arrows,decorations.markings}

\usepackage{booktabs}
\usepackage{graphicx}
\usepackage{mathtools}
\usepackage{bbm}
\usepackage{wrapfig}
\usepackage{floatrow}
\newfloatcommand{capbtabbox}{table}[][\FBwidth]

\usepackage{caption}
\usepackage{subcaption}

\usepackage{algorithm2e}
\makeatletter
\newcommand{\removelatexerror}{\let\@latex@error\@gobble}
\makeatother

\SetCommentSty{mycommfont}
\usepackage{indentfirst}

\newenvironment{tightlist}%
{\begin{list}{$\bullet$}{%
    \setlength{\topsep}{0in}
    \setlength{\partopsep}{0in}
    \setlength{\itemsep}{0in}
    \setlength{\parsep}{0in}
    \setlength{\leftmargin}{1.5em}
    \setlength{\rightmargin}{0in}
}
}%
{\end{list}
}

\newcommand{\secref}[1]{Section~\ref{#1}}

\newcommand{\eqnref}[1]{Equation~\ref{#1}}
\newcommand{\figref}[1]{Figure~\ref{#1}}

\newcommand{\thmref}[1]{Theorem~\ref{#1}}

\newtheorem{thm}{Theorem}

\counterwithin*{thm}{section}

\newcommand{\St}{\mathcal{S}}
\newcommand{\A}{\mathcal{A}}

\newcommand{\mdp}{{\sc mdp}}

\newcommand{\Ex}{\mathbb{E}}
\newcommand{\var}{\mathrm{Var}}

\DeclareMathOperator*{\argmax}{argmax}
\DeclarePairedDelimiterX{\infdivx}[2]{(}{)}{%
  #1\;\delimsize\|\;#2%
}
\newcommand{\kl}{D_{\text{KL}}\infdivx}

\title{Learning Compact Models for Planning\\with Exogenous Processes}

\author{
  Rohan Chitnis, Tom\'as Lozano-P\'erez\\\\
  MIT Computer Science and Artificial Intelligence Laboratory\\
  \texttt{\{ronuchit, tlp\}@mit.edu}
}

\begin{document}
\maketitle

%===============================================================================

\begin{abstract}
  We address the problem of approximate model minimization for \mdp s
  in which the state is partitioned into \emph{endogenous} and (much
  larger) \emph{exogenous} components. An exogenous state variable is
  one whose dynamics are independent of the agent's actions. We
  formalize the \emph{mask-learning problem}, in which the agent must
  choose a subset of exogenous state variables to reason about when
  planning; doing planning in such a reduced state space can often be
  significantly more efficient than planning in the full model. We
  then explore the various value functions at play within this
  setting, and describe conditions under which a policy for a reduced
  model will be optimal for the full \mdp. The analysis leads us to a
  tractable approximate algorithm that draws upon the notion of mutual
  information among exogenous state variables. We validate our
  approach in simulated robotic manipulation domains where a robot is
  placed in a busy environment, in which there are many other agents
  also interacting with the objects. Visit
  \href{http://tinyurl.com/chitnis-exogenous}{\texttt{http://tinyurl.com/chitnis-exogenous}}
  for a supplementary video.
\end{abstract}


\section{Introduction}
\label{sec:introduction}
Most aspects of the world are \emph{exogenous} to us: they are not
affected by our actions. However, though they are exogenous,
these processes (e.g., weather and traffic) will often play a major
role in the way we should choose to perform a task at hand. Despite
being faced with such a daunting space of processes that are out of
our control, humans are extremely adept at quickly reasoning about
\emph{which} aspects of this space they need to concern themselves
with, for the particular task at hand.

Consider the setting of a household robot tasked with doing laundry
inside a home. It should not get bogged down by reasoning about the
current weather or traffic situation, because these factors are
irrelevant to its task. If, instead, it were tasked with mowing the
lawn, then good decision-making would require it to reason about the
time of day and weather (so it can finish the task by sunset, say).

In this work, we address the problem of approximately solving Markov
decision processes (\mdp s), without too much loss in solution
quality, by leveraging the structure of their exogenous processes. We
begin by formalizing the \emph{mask-learning problem}. An autonomous
agent is given a generative model of an \mdp\ that is partitioned into
an endogenous state (i.e., that which can be affected by the agent's
actions) and a much higher-dimensional exogenous state. The agent must
choose a \emph{mask}, a subset of the exogenous state variables, that
yields a policy \emph{not too much worse} than a policy that would be
obtained by reasoning about the entire exogenous
state.% A larger mask will
% often yield higher returns, but at the cost of more expensive
% planning; conversely, a smaller mask will often yield lower returns
% (perhaps because the agent chose to ignore some aspect of the
% exogenous state that turned out to be quite relevant), but planning
% can be more efficient. We should also desire that the process of
% finding such a mask be significantly cheaper than just solving the
% whole \mdp.

After formalizing the mask-learning problem, we discuss how we can
leverage exogeneity to quickly learn transition models for only the
relevant variables from data. Then, we explore the various value
functions of interest within the problem, and discuss the conditions
under which a policy for a particular mask will be optimal for the
full \mdp. Our analysis lends theoretical credence to the idea that a
good mask should contain not only the exogenous state variables that
directly influence the agent's reward function, but also ones
whose dynamics are correlated with theirs. This idea leads to a tractable
approximate algorithm for the mask-learning problem that leverages the
structure of the \mdp, drawing upon the notion of mutual information
among exogenous state variables.

We experiment in simulated robotic manipulation domains where a robot
is put in a busy environment, among many other agents that also
interact with the objects. We show that 1) in small domains where we
can plan directly in the full \mdp, the masks learned by our
approximate algorithm yield competitive returns; 2) our approach
outperforms strategies that do not leverage the structure of the \mdp;
and 3) our algorithm can scale up to planning problems with large
exogenous state spaces.

\section{Related Work}
\label{sec:relatedwork}
The notion of an \emph{exogenous event}, one that is outside the
agent's control, was first introduced in the context of planning
problems by \citet{exogenous}. The work most similar to ours is that
by \citet{exogmdp1}, who also consider the problem of model
minimization in \mdp s by removing exogenous state variables. Their
formulation of an \mdp\ with exogenous state variables is similar to
ours, but the central assumption they make is that the reward
decomposes additively into an endogenous state component and an
exogenous state component. Under this assumption, the value function
of the full \mdp\ decomposes into two parts, and any policy that is
optimal for the endogenous \mdp\ is also optimal for the full \mdp. On
the other hand, we do not make this reward decomposition assumption,
and so our value function does not decompose; instead, our work
focuses on a different set of questions: 1) what are the conditions
under which an optimal policy in a given reduced model is optimal in
the full \mdp? 2) can we build an algorithm that leverages exogeneity
to efficiently (approximately) discover such a reduced model?

Model minimization of factored Markov decision processes is often
defined using the notion of \emph{stochastic
  bisimulation}~\cite{sbisim1,sbisim2}, which describes an equivalence
relation among states based upon their transition dynamics. Other
prior work in state abstraction tries to remove irrelevant state
variables in order to form reduced
models~\cite{irrelevant1,irrelevant2}. Our approach differs from these
techniques in two major ways: 1) we consider only reducing the
exogenous portion of the state, allowing us to develop algorithms
which leverage the computational benefits enjoyed by the exogeneity
assumption; 2) rather than trying to build a reduced model that is
faithful to the full \mdp, we explicitly optimize a different
objective (\eqnref{eq:masklearn}), which tries to find a reduced model
yielding high rewards in the full \mdp. Recent work in model-free
reinforcement learning has considered how to exploit exogenous events
for better sample complexity~\cite{inputdriven,contingencyaware},
whereas we tackle the problem from a model-based
perspective.% Directly optimizing
% this objective can be useful in many practical settings, where the
% final performance of the agent on a task is the primary consideration.

% This work can be viewed as implementing a particular type of attention
% mechanism: the agent must decide which aspects of the world to attend
% to in order to perform a task. Attention mechanisms are commonly used
% in visual perception~\cite{vattention1,vattention2} and natural
% language~\cite{nlattention1,nlattention2,nlattention3} tasks, as a
% prior informing the learned models about the sparsity of the
% input-output mapping. The masks we learn in this work are also a form
% of sparsification, in the sense that they yield a policy mapping only
% a subset of the full state space to the action space, although the
% learning techniques are very different.

If we view the state as a vector of features, then another perspective
on our approach is that it is a technique for \emph{feature
  selection}~\cite{featureselection} applied to \mdp s with exogenous
state variables.
%In particular, we focus on the problem of learning a
% subset of exogenous state variables for which ignoring all the other
% ones does not cause us to lose too much.
This is closely related to
the typical subset selection problem in supervised
learning~\cite{ss1,ss2,ss3,ss4}, in which a learner must
determine a subset of features that suffices for making predictions
with little loss.

\section{Problem Formulation}
\label{sec:problemsetting}
In this section, we introduce the notion of exogeneity in the context
of a Markov decision process, and use this idea to formalize the
mask-learning problem.

\subsection{Markov Decision Processes with Exogenous Variables}
\label{subsec:mdp}
We will formalize our setting as a discrete Markov decision process (\mdp) with
special structure. An \mdp\ is a tuple $(\St, \A, T, R, \gamma)$
where: $\St$ is the state space; $\A$ is the action space;
$T(s_t, a_t, s_{t+1}) = P(s_{t+1} \mid s_t, a_t)$ is the state
transition distribution with $s_t, s_{t+1} \in \St$ and $a_t \in \A$;
$R(s_t, a_t)$ is the reward function; and $\gamma$ is the discount
factor in $(0, 1]$. On each timestep, the agent chooses an action,
transitions to a new state sampled from $T$, and receives a reward as
specified by $R$. The solution to an \mdp\ is a \emph{policy} $\pi$, a
mapping from states in $\St$ to actions in $\A$, such that the
expected discounted sum of rewards over trajectories resulting from
following $\pi$, which is
$\Ex\left[\sum_{t=0}^{\infty} \gamma^t R(s_t, \pi(s_t))\right]$, is
maximized. Here, the expectation is with respect to the stochasticity
in the initial state and state transitions. The \emph{value} of a
state $s \in \St$ under policy $\pi$ is defined as the expected
discounted sum of rewards from following $\pi$, starting at state $s$:
$V_{\pi}(s) = \Ex\left[\sum_{t=0}^{\infty} \gamma^t R(s_t, \pi(s_t))
  \mid s_0 = s\right]$. The \emph{value function} for $\pi$ is the
mapping from $\St$ to $\mathbb{R}$ defined by
$V_{\pi}(s), \forall s \in \St$.

In this work, we assume that the agent is given a \emph{generative}
model of an \mdp, in the sense of~\citet{generative}. Concretely, the
agent is given the following:
\begin{tightlist}
\item Knowledge of $\St$, $\A$, and $\gamma$.
\item A black-box sampler of the transition model, which takes as
  input a state $s \in \St$ and action $a \in \A$, and returns a next
  state $s' \sim T(s, a, s')$.
\item A black-box reward function, which takes as input a state
  $s \in \St$ and action $a \in \A$, and returns the reward $R(s, a)$
  for that state-action pair.
\item A black-box sampler of an initial state $s_0$ from some
  distribution $P(s_0)$.
\end{tightlist}
We note that this assumption of having a generative model of an \mdp\
lies somewhere in between that of only receiving execution traces (as
in the typical reinforcement learning setting, assuming no ability to
reset the environment) and that of having knowledge of the full
analytical model. One can also view this assumption as saying that a
simulator is available. Generative models are a natural way to specify
a large \mdp, as it is typically easier to produce samples from a
transition model than to write out the complete next-state
distributions.

\begin{wrapfigure}{r}{0.4\columnwidth}
  \centering
  \noindent
  % \vspace{-1em}
    \includegraphics[width=\columnwidth]{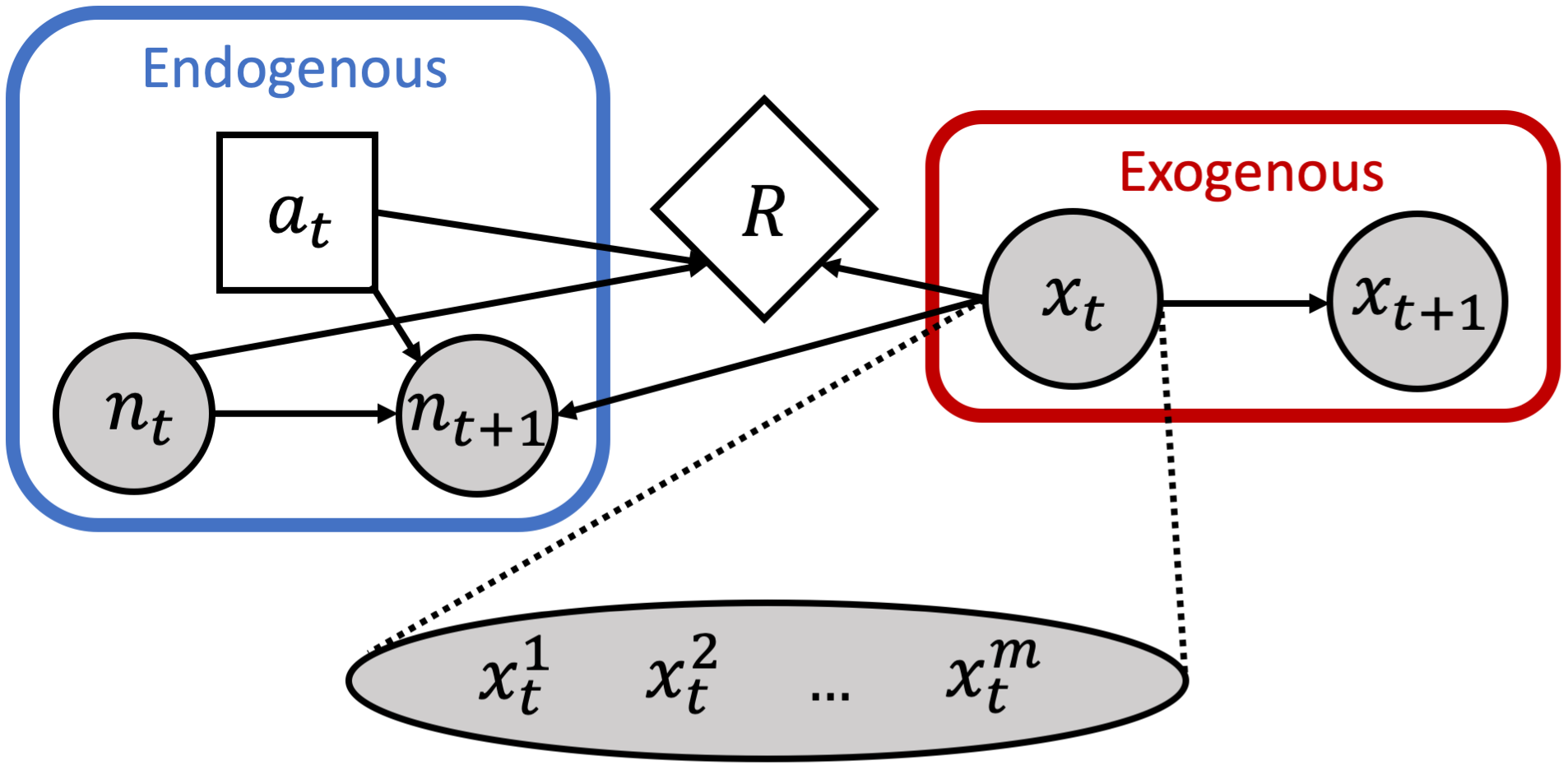}
    \caption{\small{A Markov decision process with exogenous state
        variables $x_t^1, x_t^2, \ldots, x_t^m$.}}%, unaffected by the agent's
        % actions. These variables influence both the agent's reward and
        % the transitions of the endogenous state $n_t$. This $m$ will
        % typically be very large.}}
  \label{fig:model}
\end{wrapfigure}

We assume that the state of our \mdp\ decomposes into an
\emph{endogenous} component and a (much larger) \emph{exogenous}
component, and that the agent knows this partition. Thus, we write
$s_t = \begin{bmatrix} n_t & x_t \end{bmatrix}$, where
$n_t$ is the endogenous component of the state, whose transitions the
agent can affect through its actions, and $x_t$ is the exogenous
component of the state. This assumption means that
$P(x_{t+1} \mid x_t, a_t) = P(x_{t+1} \mid x_t)$, and therefore
$P(s_{t+1} \mid s_t, a_t) = P(n_{t+1} \mid n_t, a_t, x_t) \cdot
P(x_{t+1} \mid x_t)$.

Though unaffected by the agent's actions, the exogenous state
variables influence the agent through the rewards and endogenous state
transitions. Therefore, to plan, the agent will need to reason about
future values of the $x_t$.% As discussed in \secref{sec:relatedwork},
% a similar formulation was given in~\cite{exogmdp1}, but we will not be
% making the value function decomposition assumption that was made in
% that work.

We will focus on the setting where the exogenous state $x_t$ is made
up of $m$ (not necessarily independent) state variables
$x_t^1, x_t^2, \ldots, x_t^m$, with $m$ large. The fact that $m$ is
large precludes reasoning about the entire exogenous state. We will
define the mask-learning problem (\secref{subsec:mainprob}) an
optimization problem for deciding which variables
$x^1, x^2, \ldots, x^m$ the agent should reason about.

% Note that our black-box sampler of the transition model takes as input
% $[n_t, x_t, a_t]$, and returns $[n_{t+1}, x_{t+1}]$. Therefore, due to
% the way the state transitions are factored, we can use this black-box
% sampler to gather data on the dynamics of the exogenous state by
% simply dropping the inputs $n_t$ and $a_t$, thus obtaining a sample
% from $P(x_{t+1} \mid x_t)$.

To be able to unlink the effects of each exogenous state variable on
the agent's reward, we will need to make an assumption on the form of
the reward function $R(s_t, a_t) = R(n_t, x_t, a_t)$. This assumption
is necessary for the agent to be able to reason about the effect of
dropping a particular exogenous state variable from
consideration. Specifically, we assume that
$R(n_t, x_t, a_t) = \sum_{i=1}^m R^i(n_t, x_t^i, a_t)$. In words, this
says that the reward function decomposes into a sum over the
individual effects of each exogenous state variable $x^i$. Although
this means that the computation of the agent's reward cannot include
logic based on any combinations of $x^i$, this assumption is not too
restrictive: one could always construct ``super-variables''
encompassing state variables coupled in the reward. Note that despite
this assumption, the value function might depend non-additively on the
exogenous variables.

% \subsubsection{Unobserved Exogenous Variables}
% \label{subsec:unobs}
% The \mdp\ formulation we have adopted thus far assumes that the
% exogenous variables are fully observed: the agent sees all the $x^i$
% on each timestep. In real-world problems, however, many of these
% exogenous state variables will be \emph{unobserved} during execution
% time, and the agent may have to infer their values by observing other
% correlated state variables. This setting can be formulated as a mixed
% observability \mdp\ (\momdp)~\cite{momdp} with trivial observations
% (the values of the observed variables are seen). Due to uncertainty
% over the values of the unobserved state variables, the agent will have
% to maintain a belief, a probability distribution over states;
% accordingly, the solution to a \momdp\ is a policy that maps beliefs
% to actions. \momdp s can be solved using approximate \pomdp\ solvers,
% such as online simulation-based
% methods~\cite{pomcp,despot,beliefspaceplanning} and offline
% point-based methods~\cite{sarsop,pointpomdp}.

% The methods that we present are easily adapted to the setting where
% some subset of the exogenous state variables will be unobserved during
% execution time. We will point out the necessary modifications where
% appropriate, and experimentally validate performance in a robotic
% \momdp\ domain.

\subsection{The Mask-Learning Problem}
\label{subsec:mainprob}
Our central problem of focus is that of finding a \emph{mask}, a subset of the $m$
exogenous state variables, that is ``good enough'' for
planning, in the sense that we do not lose too much by ignoring the
others. Before being able to formalize the problem, we must define
precisely what it means to plan with only a subset of the $m$
exogenous variables. Suppose we have an \mdp\ $M$ as described in
\secref{subsec:mdp}. Let
$x = \begin{bmatrix} x^1 & x^2 & \ldots & x^m \end{bmatrix}$, and
$\tilde{x} \subseteq x$ be an arbitrary subset (a \emph{mask}).

We define the \emph{reduced model} $\tilde{M}$ corresponding to mask
$\tilde{x}$ and \mdp\ $M$ as another \mdp:
\begin{tightlist}
\item $\A$ and $\gamma$ are the same as in $M$.
\item $\St$ is reduced by removing the dimensions corresponding to any
  $x^i \notin \tilde{x}$.
\item
  $P(\tilde{s}_{t+1} \mid \tilde{s}_t, a_t) = P(n_{t+1} \mid n_t, a_t,
  \tilde{x}_t) \cdot P(\tilde{x}_{t+1} \mid \tilde{x}_t)$. Here,
  $\tilde{s}_t = \begin{bmatrix} n_t &
    \tilde{x}_t \end{bmatrix}$.\footnote{The $\tilde{x}$ might, in
    actuality, not be Markov, since we are ignoring the variables
    $x \setminus \tilde{x}$. Nevertheless, this expression is
    well-formed, and estimating it from data marginalizes out these
    ignored variables. If they were not exogenous, such estimates
    would depend on the data-gathering policy (and thus could be very error-prone).}
\item
  $R(\tilde{s}_t, a_t) = R(n_t, \tilde{x}_t, a_t) = \sum_{x^i \in
    \tilde{x}} R^i(n_t, x_t^i, a_t)$. Here, we are leveraging the
  assumption that the reward function decomposes as discussed at the
  end of \secref{subsec:mdp}.
\end{tightlist}

% If $|\tilde{x}|$ is much smaller than $m$, then the state space of
% $\tilde{M}$ will be much smaller than that of $M$.
Since the agent
only has access to a generative model of the \mdp\ $M$, planning in
$\tilde{M}$ will require estimating the reduced dynamics and reward
models, $\hat{T}(\tilde{s}_t, a_t, \tilde{s}_{t+1})$ and
$\hat{R}(\tilde{s}_t, a_t)$.

Formally, the \emph{mask-learning problem} is to determine:
\begin{equation}
  \label{eq:masklearn}
  \tilde{x}^* = \argmax_{\tilde{x} \subseteq x} J(\tilde{x}) =
  \argmax_{\tilde{x} \subseteq x} \Ex\Bigg[\sum_{t=0}^{\infty}
  \gamma^t R(n_t, x_t, \tilde{\pi}(n_t, \tilde{x}_t))\Bigg] - \lambda \cdot
  \text{Cost}(\tilde{x}),
\end{equation}
where $\tilde{\pi}$ is the policy (mapping reduced states to actions)
that is obtained by planning in $\tilde{M}$. In words, we seek the
mask $\tilde{x}^*$ that yields a policy maximizing the expected total
reward \emph{accrued in the actual environment} (the complete \mdp\
$M$), minus a cost on $\tilde{x}$. Note that if $\lambda = 0$, then
the choice $\tilde{x}=x$ is always optimal, and so the $\lambda$
serves as a regularization hyperparameter that balances the expected reward with the
complexity of the considered mask. Reasonable choices of
$\text{Cost}(\tilde{x})$ include $|\tilde{x}|$ or the amount of time
needed to produce the policy $\tilde{\pi}$ corresponding to
$\tilde{x}$.

\section{Leveraging Exogeneity}
\label{sec:exog}
The agent has only a generative model of the \mdp\ $M$
(\secref{subsec:mdp}). In order to build a reduced \mdp\ $\tilde{M}$,
it must estimate
$\hat{T}(\tilde{s}_t, a_t, \tilde{s}_{t+1}) = \hat{P}(n_{t+1} \mid
n_t, a_t, \tilde{x}_t) \cdot \hat{P}(\tilde{x}_{t+1} \mid
\tilde{x}_t)$ and $\hat{R}(\tilde{s}_t, a_t)$.

% \begin{wrapfigure}{r}{0.5\columnwidth}
%   \centering
%     \noindent
%     \vspace{-1em}
%     \includegraphics[width=0.92\columnwidth]{figures/leverage.png}
%     \caption{\small{We can roll out simulations of an exogenous state
%         variable without committing to a policy, making it tractable
%         to build reduced models corresponding to only a subset of the
%         exogenous state variables. Each node in this tree represents a
%         different possible setting of values for the exogenous
%         variables, indexed by a timestep.}}
%   \label{fig:leverage}
% \end{wrapfigure}

Recall that we are considering the setting where the space of
exogenous variables is much larger than the space of endogenous
variables. At first glance, then, it seems challenging to estimate
$\hat{P}(\tilde{x}_{t+1} \mid \tilde{x}_t)$ using only the generative
model for $P(x_{t+1} \mid x_t)$. However, this estimation problem is
in fact greatly simplified due to the exogeneity of the $x^i$. To see
why, consider the typical strategy for estimating a transition model
from data: generate trajectories starting from some initial state
$x_0$, then fit a one-step prediction model to this data. Now, if the
$x^i$ were endogenous, then we would need to commit to some policy
$\pi_{\text{rollout}}$ in order to generate these trajectories, and
the reduced transition model we learn would depend heavily on this
$\pi_{\text{rollout}}$. However, since the $x^i$ are exogenous, we can
roll out simulations conditioned \emph{only on the initial state
  $x_0$, not on a policy}, and use this data to efficiently estimate the
transition model $P(\tilde{x}_{t+1} \mid \tilde{x}_t)$ of the reduced
\mdp\ $\tilde{M}$.% This idea is illustrated in \figref{fig:leverage}.
\footnote{In high dimensions, we may still need a lot of data, unless
  the state factors nicely.}

Because we can efficiently estimate the reduced model dynamics of
exogenous state variables, we can not only tractably construct the
reduced \mdp, but also allow ourselves to explore algorithms that
depend heavily on estimating these variables' dynamics, as we will do
in \secref{subsec:mainalgo}.

\section{Algorithms for Mask-Learning}
\label{sec:algos}
In this section, we explore some value functions induced by the
mask-learning problem, and use this analysis to develop a tractable
but effective approximate algorithm for finding good masks.

\subsection{Preliminaries}
\label{subsec:prelim}
Observe that the expectation in the objective $J(\tilde{x})$
(\eqnref{eq:masklearn}) is the value $V_{\tilde{\pi}}(s_0)$ of an
initial state under the policy $\tilde{\pi}$, corresponding to mask
$\tilde{x}$. Because the full \mdp\ $M$ is very large, computing this
value (the expected discounted sum of rewards) exactly will not be
possible. Instead, we can use rollouts of $\tilde{\pi}$ to produce an
estimate $\hat{V}_{\tilde{\pi}}$, which in turns yields an estimate of
the objective, $\hat{J}(\tilde{x})$:

\begin{algorithm}[H]
  \SetAlgoNoEnd
  \DontPrintSemicolon
  \SetKwFunction{algo}{algo}\SetKwFunction{proc}{proc}
  \SetKwProg{myalg}{Algorithm}{}{}
  \SetKwProg{myproc}{Procedure}{}{}
  \SetKw{Continue}{continue}
  \SetKw{Break}{break}
  \SetKw{Return}{return}
  \myproc{\textsc{Estimate-Objective}$(M, \tilde{x}, n_{\text{rollouts}})$}{
    \nl Construct estimated reduced \mdp\ $\tilde{M}$ defined by mask $\tilde{x}$ and full \mdp\ $M$.\;
    \nl Solve $\tilde{M}$ to get policy $\tilde{\pi}$.\;
    \nl \For{$i = 1, 2, \ldots, n_{\text{rollouts}}$}{
        \nl Execute $\tilde{\pi}$ in the full \mdp\ $M$, obtain total discounted returns $r_i$.
    }
    \nl $\hat{J}(\tilde{x}) = \frac{1}{n_{\text{rollouts}}}\sum_i r_i - \lambda \cdot \text{Cost}(\tilde{x})$\;
  }
\end{algorithm}

As discussed in \secref{sec:exog}, Line 1 is tractable due to the
exogeneity of the $x$ variables, which make up most of the
dimensionality of the state. With \textsc{Estimate-Objective} in hand,
we can write down some very simple strategies for solving the
mask-learning problem:
\begin{tightlist}
\item \textsc{Mask-Learning-Brute-Force}: Evaluate
  $\hat{J}(\tilde{x})$ for every possible mask
  $\tilde{x} \subseteq x$. Return the highest-scoring mask.
\item \textsc{Mask-Learning-Greedy}: Start with an empty mask
  $\tilde{x}$. While $\hat{J}(\tilde{x})$ increases, pick a variable
  $x^i$ at random, add it into $\tilde{x}$, and re-evaluate
  $\hat{J}(\tilde{x})$.
\end{tightlist}

% \begin{algorithm}[H]
%   \SetAlgoNoEnd
%   \DontPrintSemicolon
%   \SetKwFunction{algo}{algo}\SetKwFunction{proc}{proc}
%   \SetKwProg{myalg}{Algorithm}{}{}
%   \SetKwProg{myproc}{Subroutine}{}{}
%   \SetKw{Continue}{continue}
%   \SetKw{Break}{break}
%   \SetKw{Return}{return}
%   \myalg{\textsc{Mask-Learning-Brute-Force}$(M)$}{
%     \nl $x \gets$ the $m$ exogenous state variables in \mdp\ $M$\;
%     \nl \For{each subset $\tilde{x} \subseteq x$}{
%       \nl $\hat{J}(\tilde{x}) \gets $\textsc{Estimate-Objective}$(M, \tilde{x})$
%     }
%     \nl \Return the $\tilde{x}$ that maximizes $\hat{J}(\tilde{x})$
%   }
% \end{algorithm}

% \begin{algorithm}[H]
%   \SetAlgoNoEnd
%   \DontPrintSemicolon
%   \SetKwFunction{algo}{algo}\SetKwFunction{proc}{proc}
%   \SetKwProg{myalg}{Algorithm}{}{}
%   \SetKwProg{myproc}{Subroutine}{}{}
%   \SetKw{Continue}{continue}
%   \SetKw{Break}{break}
%   \SetKw{Return}{return}
%   \myalg{\textsc{Mask-Learning-Greedy}$(M)$}{
%     \nl $\tilde{x} \gets \varnothing$\;
%     \nl \While{$\hat{J}(\tilde{x})$ increases}{
%       \nl $i \gets$ a random index in $\{1, \ldots, m\}$\;
%       \nl $\tilde{x} = \tilde{x} \cup x^i$\;
%       \nl $\hat{J}(\tilde{x}) \gets $\textsc{Estimate-Objective}$(M, \tilde{x})$
%     }
%     \nl \Return $\tilde{x}$
%   }
% \end{algorithm}

While optimal, \textsc{Mask-Learning-Brute-Force} is of course
intractable for even medium-sized values of $m$, as it will not be
feasible to evaluate all $2^m$ possible subsets of $x$.
Unfortunately, even \textsc{Mask-Learning-Greedy} will likely be
ineffective for medium and large \mdp s, as it does not leverage the
structure of the \mdp\ whatsoever. To develop a better algorithm, we
will start by exploring the connection between the value functions of
the reduced \mdp\ $\tilde{M}$ and the full \mdp\ $M$.

\subsection{Analyzing the Value Functions of Interest}
It is illuminating to outline the various value functions at play
within our problem. We have:
\begin{tightlist}
\item $V^*(s)$: the value function under an optimal policy; unknown
  and difficult to compute exactly.
\item $V_{\tilde{\pi}}(s)$: given a mask $\tilde{x}$, the value
  function under the policy $\tilde{\pi}$; unknown and difficult to
  compute exactly. Note that
  $V^*(s) \geq V_{\tilde{\pi}}(s)\ \forall s \in \St$, by the
  definition of an optimal value function.
\item $\hat{V}_{\tilde{\pi}}(s)$: given a mask $\tilde{x}$, the
  empirical estimate of $V_{\tilde{\pi}}(s)$, which the agent can
  obtain by rolling out $\tilde{\pi}$ in the environment many times,
  as was done in the \textsc{Estimate-Objective} procedure.
\item $\tilde{V}_{\tilde{\pi}}(\tilde{s})$: given a mask $\tilde{x}$,
  the value function of policy $\tilde{\pi}$ \emph{within the reduced
    \mdp\ $\tilde{M}$}. Here, $\tilde{s}$ is the reduced form of state
  $s$ (i.e., the endogenous state $n$ concatenated with $\tilde{x}$).
\end{tightlist}

Intuitively, $\tilde{V}_{\tilde{\pi}}(\tilde{s})$ corresponds to the
expected reward that the agent \emph{believes} it will receive by
following $\tilde{\pi}$, which typically will not match
$V_{\tilde{\pi}}(s)$, the \emph{actual} expected reward. In general,
we cannot say anything about the ordering between
$\tilde{V}_{\tilde{\pi}}(\tilde{s})$ and $V_{\tilde{\pi}}(s)$. For
instance, if the mask $\tilde{x}$ ignores some negative effect in the
environment, then the agent will expect to receive \emph{higher}
reward than it actually receives during its rollouts. On the other
hand, if the mask $\tilde{x}$ ignores some positive effect in the
world, then the agent will expect to receive \emph{lower} reward than
it actually receives.

It is now natural to ask: under what conditions would an optimal
policy $\tilde{\pi}$ for the reduced \mdp\ $\tilde{M}$ also be optimal
for the full \mdp\ $M$? The following theorem describes sufficient
conditions:

\begin{thm}
  \label{thm:main}
  Consider an \mdp\ $M$ as defined in \secref{subsec:mdp}, with
  exogenous state variables
  $x = \begin{bmatrix} x^1 & x^2 & \ldots & x^m \end{bmatrix}$, and a
  mask $\tilde{x} \subseteq x$. Let $\bar{x} = x \setminus \tilde{x}$
  be the variables not included in the mask. If the following
  conditions hold: (1)
  $R^i(n_t, x_t^i, a_t) = 0\ \forall x^i \in \bar{x}$, (2)
  $P(n_{t+1} \mid n_t, a_t, x_t) = P(n_{t+1} \mid n_t, a_t,
  \tilde{x}_t)$, (3)
  $P(\tilde{x}_{t+1}, \bar{x}_{t+1} \mid \tilde{x}_t, \bar{x}_t) =
  P(\tilde{x}_{t+1} \mid \tilde{x}_t) \cdot P(\bar{x}_{t+1} \mid
  \bar{x}_t)$; then
  $\tilde{V}_{\tilde{\pi}}(\tilde{s}) = V_{\tilde{\pi}}(s)\ \forall s
  \in \St$. If $\tilde{\pi}$ is optimal for the reduced \mdp\
  $\tilde{M}$, then it must also be true that
  $\tilde{V}_{\tilde{\pi}}(\tilde{s}) = V^*(s)\ \forall s \in \St$.
\end{thm}
\emph{Proof:} See Appendix A. \qed

The conditions in \thmref{thm:main} are very intuitive: (1) all
exogenous variables \emph{not} in the mask do not influence the
agent's reward, (2) the endogenous state transitions do not depend on
variables not in the mask, and (3) the variables in the mask
transition independently of the ones not in the mask. If these
conditions hold, and we use an optimal planner for the reduced model,
then we will obtain a policy that is optimal \emph{for the full \mdp,
  not just the reduced one}. Based on these conditions, it is clear
that our mask-learning algorithm should be informed by two things: 1)
the agent's reward function and 2) the degree of correlation among the
dynamics of the various state variables. %Practically speaking, we should not
% expect to be able to find a small mask satisfying all these conditions
% (if one even exists), but we are studying this case in order to build
% intuition toward a tractable algorithm.

% It is straightforward to bound the difference between
% $V_{\tilde{\pi}}(s)$ and its empirical estimate
% $\hat{V}_{\tilde{\pi}}(s)$:

% \begin{thm}
%   \label{thm:hoeff}
%   Consider an \mdp\ with discount factor $\gamma$ and rewards in
%   $[0, R_{\max}]$. For policy $\tilde{\pi}$ and any state $s$, let
%   $V_{\tilde{\pi}}(s)$ be the value of $s$ under $\tilde{\pi}$, and
%   $\hat{V}_{\tilde{\pi}}(s)$ be the empirical value estimated as a
%   sample average over total discounted returns across $n$ independent
%   rollouts of $\tilde{\pi}$. Then, with probability at least
%   $1-2e^{-2n\lambda^2(1-\gamma)^2/R_{\max}^2}$, we have that
%   $|V_{\tilde{\pi}}(s)-\hat{V}_{\tilde{\pi}}(s)| \leq \lambda$.
% \end{thm}
% \emph{Proof:} This result appears in several sources
% (e.g.,~\cite{chang2013simulation}) and is a direct application of
% Hoeffding's inequality~\cite{hoeffding} to the random variable
% $X = \sum_{t=0}^{\infty} \gamma^t R(s_t, \tilde{\pi}(\tilde{s}_t))$,
% whose expectation is $\Ex[X] = V_{\tilde{\pi}}(s_0)$. Note that $X$ is
% bounded below by 0 and above by
% $\sum_{t=0}^{\infty} \gamma^t R_{\max} =
% \frac{R_{\max}}{1-\gamma}$. \qed

Hoeffding's inequality~\cite{hoeffding} allows us to bound the
difference between $V_{\tilde{\pi}}(s)$ and the empirical estimate
$\hat{V}_{\tilde{\pi}}(s)$. For rewards in the range $[0, R_{\max}]$,
discount factor $\gamma$, number of rollouts $n$, and policy
$\tilde{\pi}$, we have that for any state $s$,
$|V_{\tilde{\pi}}(s)-\hat{V}_{\tilde{\pi}}(s)| \leq \lambda$ with
probability at least
$1-2e^{-2n\lambda^2(1-\gamma)^2/R_{\max}^2}$. This justifies the use
of $\hat{V}_{\tilde{\pi}}(s)$ in place of $V_{\tilde{\pi}}(s)$, as was
done in the \textsc{Estimate-Objective} procedure. Next, we describe a
tractable algorithm for mask-learning that draws on \thmref{thm:main}.

\subsection{A Correlational Algorithm for Mask-Learning}
\label{subsec:mainalgo}
Of course, it is very challenging to directly search for a low-cost
mask $\tilde{x}$ that meets all the conditions of \thmref{thm:main},
if one even exists. However, we can use that intuition to develop an
approximate algorithm based upon greedy forward selection techniques
for feature selection~\cite{featureselection}, which at each iteration
add a single variable that most improves some performance metric.

\begin{algorithm}[H]
  \SetAlgoNoEnd
  \DontPrintSemicolon
  \SetKwFunction{algo}{algo}\SetKwFunction{proc}{proc}
  \SetKwProg{myalg}{Algorithm}{}{}
  \SetKwProg{myproc}{Subroutine}{}{}
  \SetKw{Continue}{continue}
  \SetKw{Break}{break}
  \SetKw{Return}{return}
  \SetKw{Emit}{emit}
  \myalg{\textsc{Mask-Learning-Correlational}$(M, \tau_{\text{correl}}, \tau_{\text{variance}}, n_1, n_2)$}{
    \nl $\tilde{x} \gets$ \textsc{Estimate-Reward-Variables}$(M, \tau_{\text{variance}}, n_1, n_2)$ \tcp*{\footnotesize Initial mask for \mdp\ $M$.}
    \nl \While(\tcp*[f]{\footnotesize Iteratively add to initial mask.}){$\hat{J}(\tilde{x})$ increases}{
      \nl $x^i = \argmax_{x^j \notin \tilde{x}} \kl{\hat{T}_{\tilde{s}, x^j}}{\hat{T}_{\tilde{s}} \otimes \hat{T}_{x^j}}$ \tcp*{\footnotesize Measure mutual informations.}
      \nl \If{$\kl{\hat{T}_{\tilde{s}, x^i}}{\hat{T}_{\tilde{s}} \otimes \hat{T}_{x^i}} < \tau_{\text{correl}}$}{
        \Break \tcp*{\footnotesize All mutual informations below threshold, terminate.}
      }
      \nl $\tilde{x} = \tilde{x} \cup x^i$\;
      \nl $\hat{J}(\tilde{x}) \gets $\textsc{Estimate-Objective}$(M, \tilde{x})$
    }
    \nl \Return $\tilde{x}$
  }\;
  \myproc{\textsc{Estimate-Reward-Variables}$(M, \tau_{\text{variance}}, n_1, n_2)$}{
    \nl \For{each variable $x^i \in x$}{
      \nl \For{$n_1$ random samples of endog. state, exog. variables $x \setminus x^i$, action $a$}{
        \nl Randomly sample $n_2$ settings of variable $x^i$, construct full states $s_1, s_2, \ldots, s_{n_2}$.\;
        \nl Compute $\sigma_j^2 \gets \var(\{R(s_1, a), \ldots, R(s_{n_2}, a)\})$.
      }
      \nl \If(\tcp*[f]{\footnotesize Mean variance above threshold, accept.}){$\frac{1}{n_1}\sum_{j=1}^{n_1}\sigma_j^2 > \tau_{\text{variance}}$}{
        \nl \Emit $x^i$
      }
    }
  }
\end{algorithm}

\textbf{Algorithm description.} In line with Condition (1) of
\thmref{thm:main}, we begin by estimating the set of exogenous state
variables relevant to the agent's reward function, which we can do by
leveraging the generative model of the \mdp. As shown by the
\textsc{Estimate-Reward-Variables} subroutine, the idea is to compute,
for each variable $x^i$, the average amount of variance in the reward
across different values of $x^i$ when the remainder of the state is
held fixed, and threshold this at $\tau_{\text{variance}}$. We define
our initial mask to be this set. The efficiency-accuracy tradeoff of
this subroutine can be controlled by tuning the number of samples,
$n_1$ and $n_2$. In practice, we can make a further improvement by
heuristically biasing the sampling to favor states that are more
likely to be encountered by the agent (such heuristics can be
computed, for instance, by planning in a relaxed version of the
problem).

Then, we employ an iterative procedure to approximately build toward
Conditions (2) and (3): that the variables not in the mask transition
independently of those in the mask and the endogenous state. To do so,
we greedily add variables into the mask based upon the mutual
information between the empirical transition model of the reduced
state and that of each remaining variable. Intuitively, this
quantitatively measures: ``\emph{How much better would we be able to
  predict the dynamics of $\tilde{s}$ if $\tilde{s}$ included variable
  $x^i$, versus if it didn't?}''  This mutual information will be 0
if $x^i$ transitions independently of all variables in
$\tilde{s}$. %Note that if Conditions (2) and (3) are
% satisfied, then all estimated mutual informations should be below the
% threshold $\tau_{\text{correl}}$, and we will terminate.
To calculate the mutual information between $\tilde{s}$ and variable
$x^i$, we must first learn the empirical transition models
$\hat{T}_{\tilde{s}, x^i}$, $\hat{T}_{\tilde{s}}$, and $\hat{T}_{x^i}$
from data. The exogeneity of most of the state is critical here: not only
does it make learning these models much more efficient, but also,
without exogeneity, we could not be sure whether two variables
actually transition independently of each other or we just happened
to follow a data-gathering policy that led us to believe so.

\section{Experiments}
\label{sec:experiments}
Our experiments are designed to answer the following questions: (1) In
small domains, are the masks learned by our algorithm competitive with
the optimal masks? (2) Quantitatively, how well do the learned masks
perform in large, complicated domains?  (3) Qualitatively, do the
learned masks properly reflect different goals given to the robot? (4)
What are the limitations of our approach?

We experiment in simulated robotic manipulation domains in which a
robot is placed in a busy environment with objects on tables, among
many other agents that are also interacting with objects. The robot is
rewarded for navigating to a given goal object (which changes on each
episode) and penalized for crashing into other agents. The exogenous
variables are the states of the other agents\footnote{It is true that
  typically, the other agents will react to the robot's actions, but
  it is well worth making the exogeneity assumption in these
  complicated domains we are considering. This assumption is akin to
  how we treat traffic patterns as exogenous to ourselves, even though
  technically we can slightly affect them.}, a binary-valued occupancy
grid discretizing the environment, the object placements on tables,
and information specifying the goal. We plan using value iteration
with a timeout of 60 seconds.  Empirical value estimates are computed
as averages across 500 independent rollouts. Each result reports an
average across 50 independent trials. We regularize masks by setting
$\text{Cost}(\tilde{x}) = |\tilde{x}|$; our initial experimentation
suggested that other mask choices, such as planning time, perform
similarly.

\subsection{In small domains, are our learned masks competitive with
  the optimal masks?}

\begin{wrapfigure}{r}{0.5\columnwidth}
  \centering
  \vspace{-1.5em}
  \noindent
  \includegraphics[width=0.49\columnwidth]{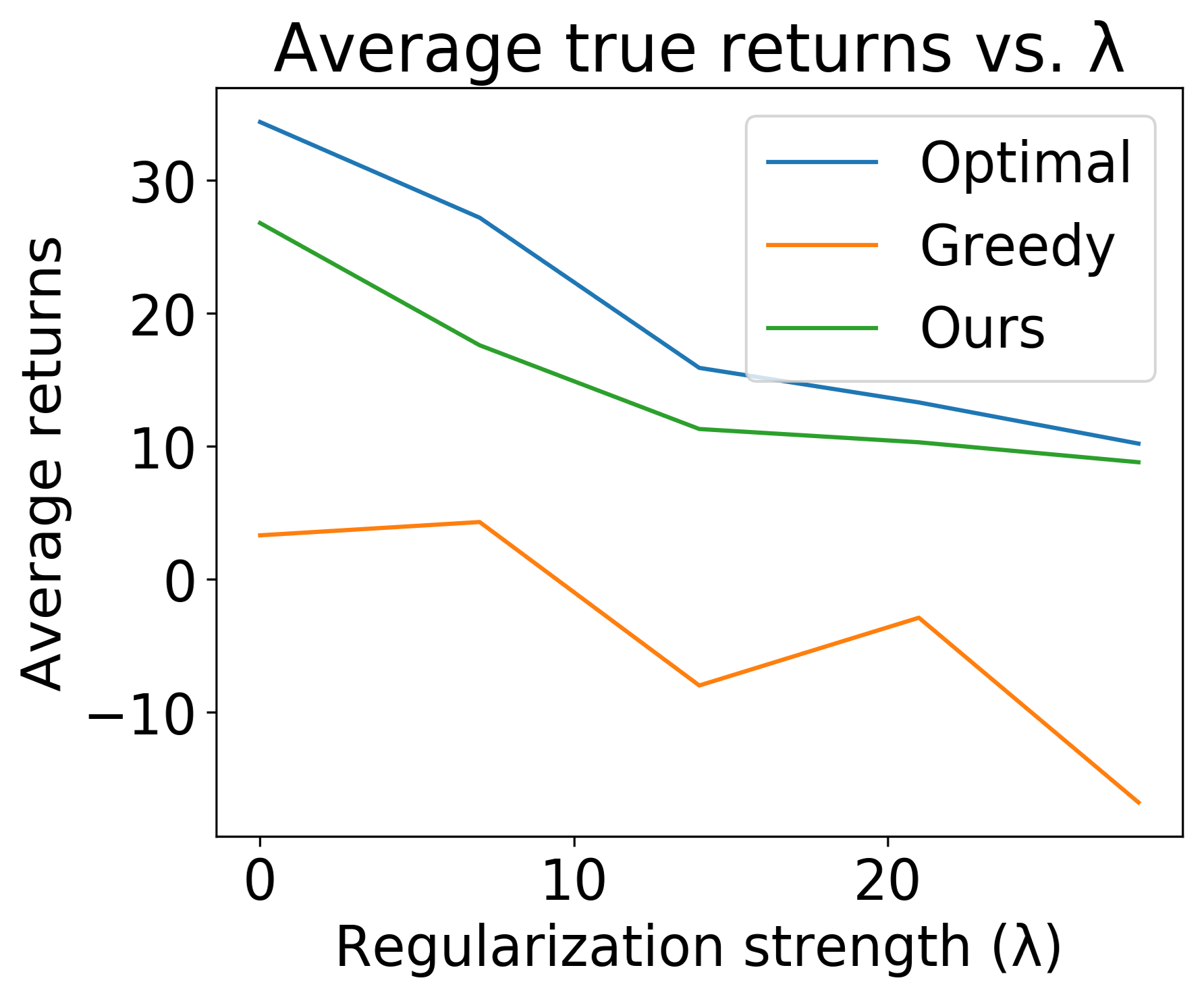}
  \includegraphics[width=0.49\columnwidth]{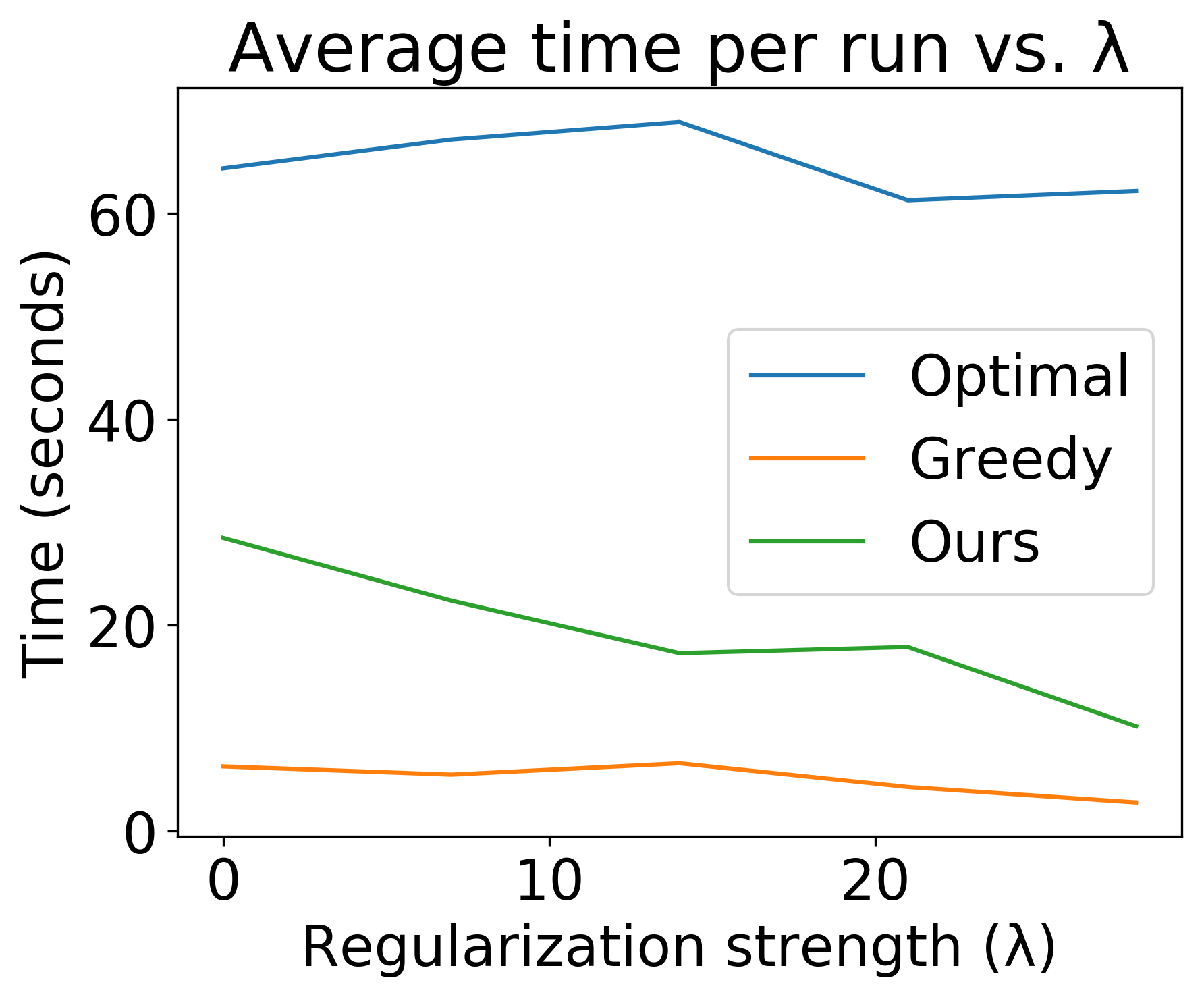}
\end{wrapfigure}
Finding the optimal mask $\tilde{x}^*$ is only
possible in small models. Therefore, to answer this question we
constructed a small gridworld representation of our experimental
domain, with only $\sim$600 states and 5 exogenous variables, in which
we can plan exactly. We find $\tilde{x}^*$ using the
\textsc{Mask-Learning-Brute-Force} strategy given in
\secref{subsec:prelim}. We compare the optimal mask with both 1)
\emph{Ours}, the mask returned by our algorithm in
\secref{subsec:mainalgo}; and 2) \emph{Greedy}, the mode mask chosen
across 10 independent trials of the \textsc{Mask-Learning-Greedy}
strategy given in \secref{subsec:prelim}.

\textbf{Discussion.} For higher values of $\lambda$ especially, our
algorithm performs on par with the optimal brute-force strategy, which
is only viable in small domains. The gap widens as $\lambda$ decreases
because as the optimal mask gets larger, it becomes harder to find
using a forward selection strategy such as ours. In practical
settings, one should typically set $\lambda$ quite high so that
smaller masks are preferred, as these will yield the most compact
reduced models. Meanwhile, the greedy strategy does not perform well
because it disregards the structure of the \mdp\ and the conditions of
\thmref{thm:main}. We also observe that our algorithm takes
significantly less time than the optimal brute-force strategy; we
should expect this gap to widen further in larger domains. Our
algorithm's stopping condition is that the score function begins to
decrease, and so it tends to terminate more quickly for higher values
of $\lambda$, as smaller masks become increasingly preferred.

\subsection{Quantitatively, how well do the learned masks perform in
  large, complicated domains?}
To answer this question, we consider two large domains:
\emph{Factory}, in which a robot must fulfill manipulation tasks being
issued in a stream; and \emph{Crowd}, in which a robot must navigate
to target objects among a crowd of agents executing their own fixed
stochastic policies. Even after discretization, these domains have
$10^{11}$ and $10^{80}$ states respectively; \emph{Factory} has 22
exogenous variables and \emph{Crowd} has 124. In either domain, both
planning exactly in the full \mdp\ and searching for $\tilde{x}^*$ by
brute force are prohibitively expensive.  All results use
$n_{\text{rollouts}} = 500$, $\tau_{\text{correl}} = 10^{-5}$,
$\tau_{\text{variance}} = 0$, $n_1 = 250$, and $n_2 = 5$. We compare
our algorithm with the \emph{Greedy} one described earlier, which
disregards the structure of the \mdp. We also compare to only running our algorithm's
first phase, which chooses the mask to be the estimated set of variables directly influencing
the reward.

\begin{figure}[h]
  \begin{floatrow}
    \hspace{-7em}
    \capbtabbox{%
      \resizebox{0.85\columnwidth}{!}{
  \begin{tabular}{c|c|c|c|c}
    \toprule[1.5pt]
    \textbf{Algorithm} & \textbf{Domain} & \textbf{Average True Returns} & \textbf{Time / Run (sec)}\\
    \midrule[2pt]
    Greedy & Factory & 0 & 13.4\\
    \midrule
    Ours (first phase) & Factory & 186 & ---\\
    \midrule
    Ours (full) & Factory & \textbf{226} & 53.8\\
    \midrule[1.5pt]
    Greedy + heuristics & Crowd & 188 & 43.7\\
    \midrule
    Ours (first phase) & Crowd & 260 & ---\\
    \midrule
    Ours (full) & Crowd & \textbf{545} & 123.5\\
    \bottomrule[1.5pt]
  \end{tabular}}
}{%
  \caption*{}%
}
\hspace{-2em}
\includegraphics[width=0.4\columnwidth]{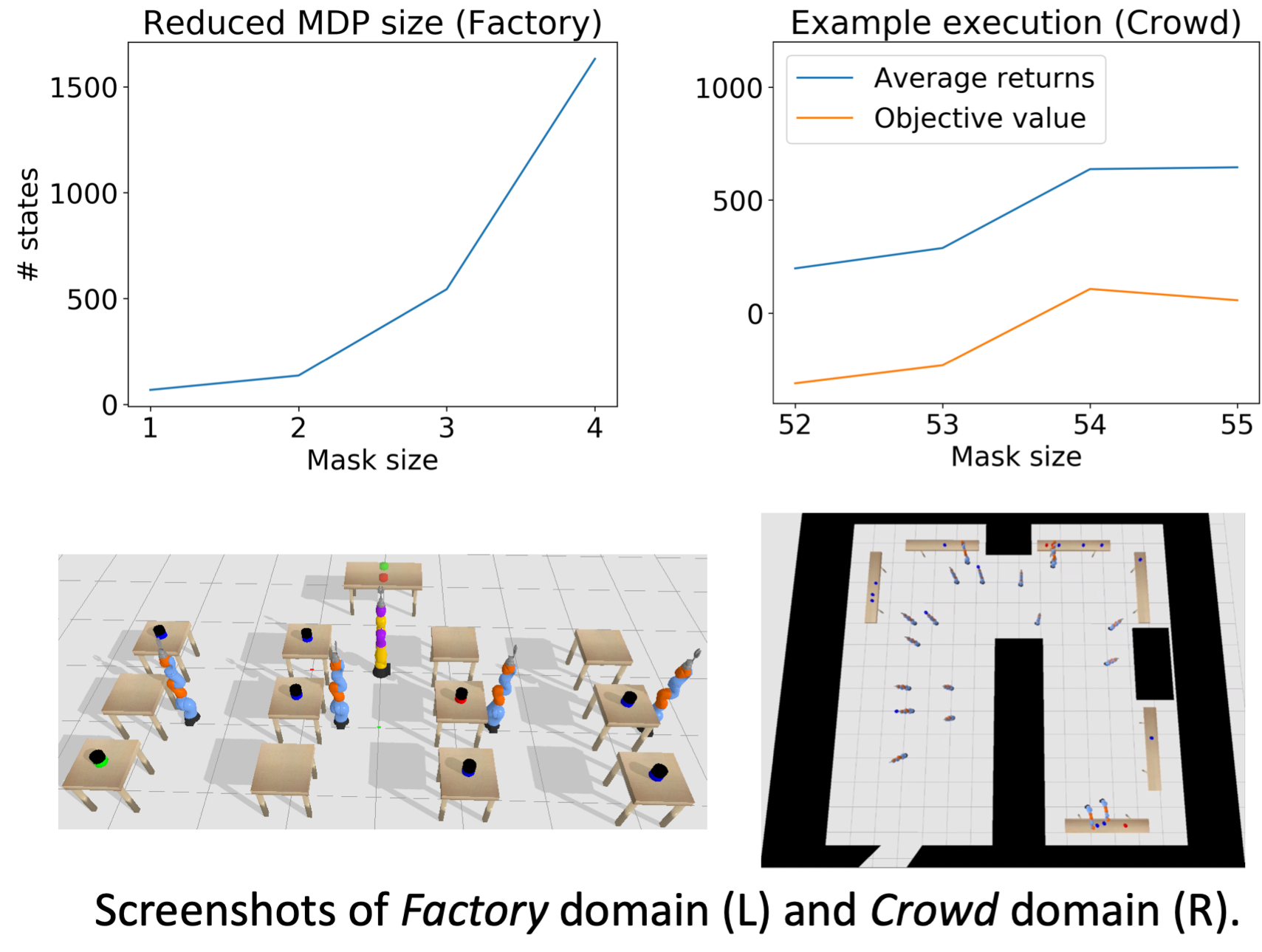}
\end{floatrow}
\end{figure}

\textbf{Discussion.} In the \emph{Factory} domain, the baseline greedy
algorithm did not succeed even once at the task. The reason is that in
this domain, several exogenous variables directly influence the reward
function, but the greedy algorithm starts with an empty mask and only
adds one variable at a time, and so cannot detect the improvement
arising from adding in several variables at once. To give the baseline
a fair chance, we initialized it by hand to a better mask for the
\emph{Crowd} domain, which is why it sees success there (\emph{Greedy
  + heuristics} in the table). The results suggest that our algorithm,
explicitly framed around \emph{all} conditions of \thmref{thm:main},
performs well. The graph of the example execution in the \emph{Crowd}
domain shows that adding a 55th variable to the mask yields a decrease
in the estimated objective even though the average returns slightly
increase, due to the regularizer $|\tilde{x}|$.

\subsection{Qualitatively, do the learned masks properly reflect
  different goals given to the robot?}
\emph{Please see the
  \href{http://tinyurl.com/chitnis-exogenous}{supplementary video} for
  more qualitative results, examples, and explanations.}

\begin{wrapfigure}{r}{0.5\columnwidth}
  \centering
  \vspace{-1em}
  \noindent
  \includegraphics[width=0.32\columnwidth]{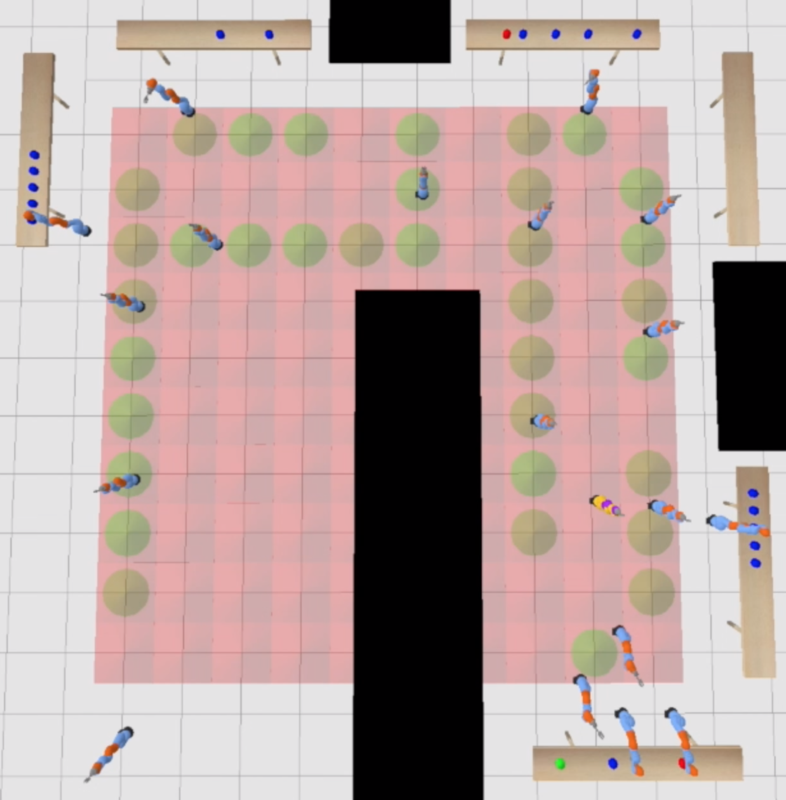}
  \includegraphics[width=0.32\columnwidth]{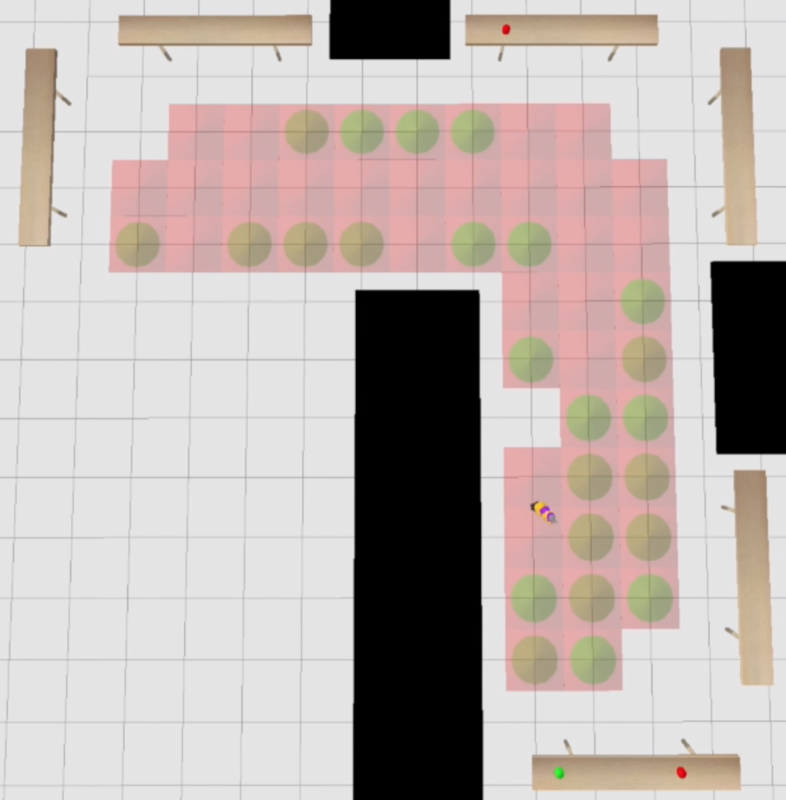}
  \includegraphics[width=0.32\columnwidth]{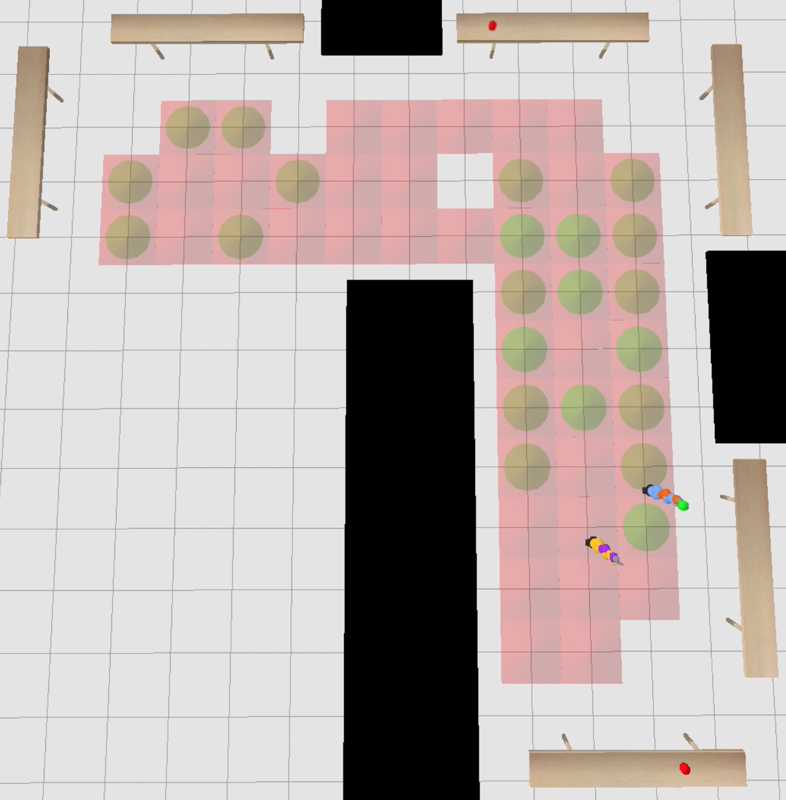}
  \caption{\small{\emph{Left}: Example of the full world. We control
      the singular gold-and-purple robot in the environment; the
      others follow fixed policies and are exogenous. \emph{Medium,
        Right}: Examples of reduced models learned by the robot for
      Goals (1) and (2). The red squares on the ground describe which
      locations, within a discretization of the environment, the robot
      has learned to consider within its masked occupancy grid, while
      green circles denote currently occupied locations. Because all
      goals require manipulating objects on the tables, the robot
      recognizes that it does not need to consider occupancies in the
      lower-left quarter of the environment. For Goal (1), in which
      the other agents cannot manipulate the objects, the robot
      recognizes that it does not need to consider the states of any
      other agents. For Goal (2), the robot considers one of the other
      agents (here, holding the green object) within its reduced
      model, since this helps it better predict the dynamics of the
      objects.}}
  \label{fig:goals}
\end{wrapfigure}

An important characteristic of our algorithm is its ability to learn
different masks based on what the goal is; we illustrate this concept
with an example. Let us explore the masks resulting from two different
goals in the \emph{Crowd} domain: Goal (1) is for the robot to
navigate to an object that cannot be moved by any of the other agents,
and Goal (2) is for the robot to navigate to an object that is
manipulable by the other agents. In either case, the variables that
directly affect the reward function are the object placements on the
tables (which tell the robot where it must navigate to) and the
occupancy grid (which helps the robot avoid crashing). However, for
Goal (2), there is another variable that is important to consider: the
states of any other agents that can manipulate the objects. This
desideratum gets captured by the second phase of our algorithm:
reasoning about the states of the other agents will allow the robot to
better predict the dynamics of the object placements, enabling it to
succeed at its task more efficiently and earn higher rewards. See
\figref{fig:goals} for a visualization of this concept in our
experimental domain, simulated using pybullet~\cite{pybullet}.

\textbf{Discussion.} In a real-world setting, all of the exogenous
variables could potentially be relevant to solving \emph{some}
problem, but typically only a small subset will be relevant to a
\emph{particular} problem. Under this lens, our method gives a way of
deriving option policies in lower-dimensional subspaces.

\vspace{-0.1em}
\subsection{What are the limitations of our approach?}
Our experimentation revealed some limitations of our approach that are
valuable to discuss. If the domains of the exogenous variables are
large (or continuous), then it is expensive to compute the necessary
mutual information quantities. To remedy this issue, one could turn to
techniques for estimating mutual information, such as
MINE~\cite{mine}. Another limitation is that the algorithm as
presented greedily adds one variable to the mask at a time, after the
initial mask is built. In some settings, it can be useful to instead
search over groups of variables to add in all at once, since
these may contain information for better predicting dynamics that is
not present in any single variable.

\section{Conclusion and Future Work}
\label{sec:conclusion}
We have formalized and given a tractable algorithm for the
\emph{mask-learning problem}, in which an agent must choose a subset
of the exogenous state variables of an \mdp\ to reason about when
planning. An important avenue for future work is to remove the
assumption that the agent knows the partition of endogenous versus
exogenous aspects of the state. An interesting fact to ponder is that
the agent can actually \emph{control} this partition by choosing its
actions appropriately. Thus, the agent can commit to a particular
choice of exogenous variables in the world, and plan under the
constraint of never influencing these variables. Another avenue for
future work is to develop an incremental, real-time version of the
algorithm, necessary in settings where the agent's task constantly
changes.

\clearpage
% The acknowledgments are automatically included only in the final version of the paper.
\acknowledgments{We gratefully acknowledge support from NSF grants
  1523767 and 1723381; from AFOSR grant FA9550-17-1-0165; from ONR
  grant N00014-18-1-2847; from Honda Research; and from the
  MIT-Sensetime Alliance on AI. Rohan is supported by an NSF Graduate
  Research Fellowship. Any opinions, findings, and conclusions or
  recommendations expressed in this material are those of the authors
  and do not necessarily reflect the views of our sponsors.}

%===============================================================================

\bibliography{references}  % .bib
\newpage
\section*{Appendix A: Proof of \thmref{thm:main}}
\begin{thm}
  Consider an \mdp\ $M$ as defined in \secref{subsec:mdp}, with
  exogenous state variables
  $x = \begin{bmatrix} x^1 & x^2 & \ldots & x^m \end{bmatrix}$, and a
  mask $\tilde{x} \subseteq x$. Let $\bar{x} = x \setminus \tilde{x}$
  be the variables not included in the mask. If the following
  conditions hold: (1)
  $R^i(n_t, x_t^i, a_t) = 0\ \forall x^i \in \bar{x}$, (2)
  $P(n_{t+1} \mid n_t, a_t, x_t) = P(n_{t+1} \mid n_t, a_t,
  \tilde{x}_t)$, (3)
  $P(\tilde{x}_{t+1}, \bar{x}_{t+1} \mid \tilde{x}_t, \bar{x}_t) =
  P(\tilde{x}_{t+1} \mid \tilde{x}_t) \cdot P(\bar{x}_{t+1} \mid
  \bar{x}_t)$; then
  $\tilde{V}_{\tilde{\pi}}(\tilde{s}) = V_{\tilde{\pi}}(s)\ \forall s
  \in \St$. If $\tilde{\pi}$ is optimal for the reduced \mdp\
  $\tilde{M}$, then it must also be true that
  $\tilde{V}_{\tilde{\pi}}(\tilde{s}) = V^*(s)\ \forall s \in \St$.
\end{thm}

\emph{Proof:} Consider an arbitrary state $s \in \St$, with
corresponding reduced state $\tilde{s}$. We begin by showing that
under the stated conditions,
$\tilde{V}_{\tilde{\pi}}(\tilde{s}) = V_{\tilde{\pi}}(s)$. The
recursive form of these value functions is:
\begin{equation*}
  V_{\tilde{\pi}}(s) = R(s, \tilde{\pi}(\tilde{s})) + \gamma \sum_{s'} P(s' \mid s, \tilde{\pi}(\tilde{s})) \cdot V_{\tilde{\pi}}(s'),
\end{equation*}
\begin{equation*}
  \tilde{V}_{\tilde{\pi}}(\tilde{s}) = R(\tilde{s}, \tilde{\pi}(\tilde{s})) + \gamma \sum_{\tilde{s'}} P(\tilde{s'} \mid \tilde{s}, \tilde{\pi}(\tilde{s})) \cdot \tilde{V}_{\tilde{\pi}}(\tilde{s'}).
\end{equation*}

Now, consider an iterative procedure for obtaining these value
functions, which repeatedly applies the above equations starting from
$V^0_{\tilde{\pi}}(s) = \tilde{V}^0_{\tilde{\pi}}(\tilde{s}) = 0\
\forall s \in \St$. Let the value functions at iteration $k$ be
denoted as $V^k_{\tilde{\pi}}(s)$ and
$\tilde{V}^k_{\tilde{\pi}}(\tilde{s})$. We will show by induction on
$k$ that
$V^k_{\tilde{\pi}}(s) = \tilde{V}^k_{\tilde{\pi}}(\tilde{s})\ \forall
k$.

The base case is immediate. Suppose
$V^k_{\tilde{\pi}}(s) = \tilde{V}^k_{\tilde{\pi}}(\tilde{s})\ \forall
s \in \St$, for some value of $k$. We compute:
\begin{align*}
  V^{k+1}_{\tilde{\pi}}(s) &= R(s, \tilde{\pi}(\tilde{s})) + \gamma \sum_{s'} P(s' \mid s, \tilde{\pi}(\tilde{s})) \cdot V^k_{\tilde{\pi}}(s')
  \\&= \sum_{i=1}^m R^i(n, x^i, \tilde{\pi}(\tilde{s})) + \gamma \sum_{s'} P(n' \mid n, \tilde{\pi}(\tilde{s}), x) \cdot P(x' \mid x) \cdot V^k_{\tilde{\pi}}(s') && \text{\color{blue} Model assumptions.}
  \\&= \sum_{x^i \in \tilde{x}} R^i(n, x^i, \tilde{\pi}(\tilde{s})) + \gamma \sum_{s'} P(n' \mid n, \tilde{\pi}(\tilde{s}), x) \cdot P(x' \mid x) \cdot V^k_{\tilde{\pi}}(s') && \text{\color{blue} Condition (1).}
  \\&= R(\tilde{s}, \tilde{\pi}(\tilde{s})) + \gamma \sum_{s'} P(n' \mid n, \tilde{\pi}(\tilde{s}), x) \cdot P(x' \mid x) \cdot V^k_{\tilde{\pi}}(s') && \text{\color{blue} Defn. of reduced $R$.}
  \\&= R(\tilde{s}, \tilde{\pi}(\tilde{s})) + \gamma \sum_{s'} P(n' \mid n, \tilde{\pi}(\tilde{s}), \tilde{x}) \cdot P(x' \mid x) \cdot V^k_{\tilde{\pi}}(s') && \text{\color{blue} Condition (2).}
  \\&= R(\tilde{s}, \tilde{\pi}(\tilde{s})) + \gamma \sum_{n',\tilde{x'}} \sum_{\bar{x'}} P(n' \mid n, \tilde{\pi}(\tilde{s}), \tilde{x}) \cdot P(\tilde{x'}, \bar{x'} \mid \tilde{x}, \bar{x}) \cdot V^k_{\tilde{\pi}}(s') && \text{\color{blue} Split up sum.}
  \\&= R(\tilde{s}, \tilde{\pi}(\tilde{s})) + \gamma \sum_{n',\tilde{x'}} \sum_{\bar{x'}} P(n' \mid n, \tilde{\pi}(\tilde{s}), \tilde{x}) \cdot P(\tilde{x'} \mid \tilde{x}) \cdot P(\bar{x'} \mid \bar{x}) \cdot V^k_{\tilde{\pi}}(s') && \text{\color{blue} Condition (3).}
  \\&= R(\tilde{s}, \tilde{\pi}(\tilde{s})) + \gamma \sum_{n',\tilde{x'}} \sum_{\bar{x'}} P(n' \mid n, \tilde{\pi}(\tilde{s}), \tilde{x}) \cdot P(\tilde{x'} \mid \tilde{x}) \cdot P(\bar{x'} \mid \bar{x}) \cdot \tilde{V}^k_{\tilde{\pi}}(n', \tilde{x'}) && \text{\color{blue} Inductive assumption.}
  \\&= R(\tilde{s}, \tilde{\pi}(\tilde{s})) + \gamma \sum_{n',\tilde{x'}} P(n' \mid n, \tilde{\pi}(\tilde{s}), \tilde{x}) \cdot P(\tilde{x'} \mid \tilde{x}) \cdot \tilde{V}^k_{\tilde{\pi}}(n', \tilde{x'}) \cdot \cancel{\sum_{\bar{x'}} P(\bar{x'} \mid \bar{x})} && \text{\color{blue} Rearrange.}
  \\&= R(\tilde{s}, \tilde{\pi}(\tilde{s})) + \gamma \sum_{\tilde{s'}} P(\tilde{s'} \mid \tilde{s}, \tilde{\pi}(\tilde{s})) \cdot \tilde{V}^k_{\tilde{\pi}}(\tilde{s'}) && \text{\color{blue} Defn. of reduced $T$.}
  \\&= \tilde{V}^{k+1}_{\tilde{\pi}}(\tilde{s}).
\end{align*}

We have shown that
$V^k_{\tilde{\pi}}(s) = \tilde{V}^k_{\tilde{\pi}}(\tilde{s})\ \forall
k$. By standard arguments (e.g., in~\cite{mdp}), this iterative procedure
converges to the true $V_{\tilde{\pi}}(s)$ and
$\tilde{V}_{\tilde{\pi}}(\tilde{s})$ respectively. Therefore, we have
that $\tilde{V}_{\tilde{\pi}}(\tilde{s}) = V_{\tilde{\pi}}(s)$.

Now, if $\tilde{\pi}$ is optimal for $\tilde{M}$, then it is optimal
for the full \mdp\ $M$ as well. This is because Condition (1) assures
us that the variables not considered in the mask do not affect the
reward, implying that $\tilde{\pi}$ optimizes the expected reward in
not just $\tilde{M}$, but also $M$. Therefore, under this assumption,
we have that
$\tilde{V}_{\tilde{\pi}}(\tilde{s}) = V_{\tilde{\pi}}(s) = V^*(s)\
\forall s \in \St$. \qed

\end{document}